# Contextual Multi-Scale Region Convolutional 3D Network for Activity Detection

Yancheng Bai[1], Huijuan Xu[2], Kate Saenko[2], Bernard Ghanem[1]
[1]KAUST, Saudi Arabia, [2]Boston University, USA, [3]Institute of Software, CAS, China
{yancheng.bai, bernard.ghanem}@kaust.edu.sa, {hxu, saenko}@bu.edu

## Abstract

*Activity detection is a fundamental problem in computer vision. Detecting activities of different temporal scales is particularly challenging. In this paper, we propose the contextual multi-scale region convolutional 3D network (CMS-RC3D) for activity detection. To deal with the inherent temporal scale variability of activity instances, the temporal feature pyramid is used to represent activities of different temporal scales. On each level of the temporal feature pyramid, an activity proposal detector and an activity classifier are learned to detect activities of specific temporal scales. Temporal contextual information is fused into activity classifiers for better recognition. More importantly, the entire model at all levels can be trained end-to-end. Our CMS-RC3D detector can deal with activities at all temporal scale ranges with only a single pass through the backbone network. We test our detector on two public activity detection benchmarks, THUMOS14 and ActivityNet. Extensive experiments show that the proposed CMS-RC3D detector outperforms state-of-the-art methods on THUMOS14 by a substantial margin and achieves comparable results on ActivityNet despite using a shallow feature extractor.*

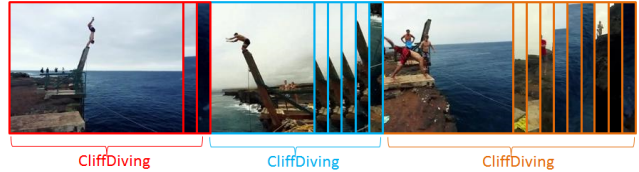

Figure 1. In unconstrained settings, activity instances in videos have varying durations. However, most of prior activity detectors deploy feature maps of the fixed temporal resolution for activity detection. This creates an inconsistency between the inherent variability in temporal length of activity instances and the fixed temporal resolution of features, thus, leading inferior performance.

## 1. Introduction

Temporal activity detection has attracted a lot of attention in computer vision in recent years due to its many applications, including content based video retrieval for web search engines, activity monitoring in video surveillance, etc. Activity detection in untrimmed videos is a challenging problem, since instances of an activity can happen at arbitrary times in arbitrarily long videos. Many works have been proposed to solve this problem and much progress has been made [25, 24, 3, 33, 8, 37]. However, how to detect temporal activity boundaries in untrimmed videos precisely is still an open question.

Inspired by the great success of R-CNN [10] and Faster RCNN [22] in object detection, most existing state-of-the-art activity detection approaches [25, 3, 33, 37] deal with this problem as detection by classification. In most cases, a subset of proposals are generated via sliding window or proposal methods and then a classifier is used to classify these temporal segments as a specific activity category. Most approaches suffer from the following major drawbacks. 1) When detecting activity instances of multiple temporal scales, they rely on either hand-crafted feature maps, or deep convolutional (*conv*) feature maps like VGG [27], Inception [29], ResNet [11] or C3D [30] with a fixed temporal resolution. However, as shown in Figure 1, activity instances of the same category in the same video have different temporal scales (durations). Thus, there is an inconsistency between the inherent temporal scale variability of activity instances and the fixed temporal resolution of the feature maps, which might degrade the detection performance. Clearly, we can deal with this problem by observing video segments at multiple temporal scales like multi-scale sliding windows, however, this increases inference time. 2) With fixed temporal resolution of the *conv* feature maps, contextual information, which has been demonstrated to be very effective in describing activities and boosting activity classification performance [19, 12, 34], is not fully exploited.

Following the detection by classification framework, we propose the contextual multi-scale (duration) region convolutional 3D network (CMS-RC3D) for activity detection in videos. To deal with activities of various temporal scales, we create the multi-scale temporal feature pyramid to represent activities. One model is trained on each level of the temporal feature pyramid to detect activity instances within specific duration ranges. Each model has its own *conv* feature maps of different temporal resolutions, which can handle activities of different temporal scales better than using a single scale approach. In doing so, contextual information is fused when recognizing activities, thus, enabling our proposed detector to achieve state-of-the-art performance.

**Contributions.** This paper makes three main contributions. **(1)** A new multi-scale region convolutional 3D network architecture for activity detection is proposed, where each scale model is trained to handle activities of a specific temporal scale range. More importantly, the detector can detect activities in videos with a variety of temporal scales in a single pass. **(2)** Contextual information is fused to recognize activities, which can improve detection performance. **(3)** The proposed CMS-RC3D detector achieves state-of-the-art results on THUMOS14 benchmark, and gets improved results on ActivityNet compared to the original RC3D model without multi-scale and context consideration.

## 2. Related Work

### 2.1. Temporal Activity Detection

Temporal activity detection needs to locate when and which type of activity happens in untrimmed diverse videos, leading to a very challenging task. In contrast to object detection, most existing temporal activity detection approaches [25, 33, 37, 3] can be categorized into the detection by classification framework. S-CNN [25] proposes a three-stage CNN for candidate segment generation, action classification and temporal boundary refinement. To deal with the limited temporal resolution issue it faces, S-CNN needs to consider video segments of varied temporal lengths and then uniformly samples 16 frames for C3D [30] feature extraction. Recently, an end-to-end trainable network named R-C3D [33] was proposed, which learns task-dependent convolutional features by jointly optimizing proposal generation and activity classification. R-C3D utilizes a fixed temporal length feature for multi-scale activity detection, which might contribute to imprecise results when detecting activities of various durations in videos. In [37], to deal with the limited temporal resolution issue, a heuristic multi-scale grouping strategy is proposed in the structured segment network (SSN). Compared to these methods, the proposed CMS-RC3D detector constructs the temporal feature pyramid and learns a multi-scale model end-to-end to deal with activities of different temporal scales.

### 2.2. Context Modeling for Activity Detection

Many approaches have incorporated context cues to boost activity recognition performance [13, 32]. In [13], the co-occurrence between activities and a sparse number of objects has been studied. In [32], neural network are introduced to learn object, scene, and activities relationships to improve classification accuracy. However, these approaches focus on activity recognition rather than activity detection. In [3], semantic context information garnered from activity-object and activity-scene relationships is exploited to address challenges in activity detection. However, it requires an extra generic object proposal generator and classifier to prune out activity proposals and classes that are not likely in the context. As such, it inherently has the same limited temporal resolution issue as before, requires more processing and semantic information, and is not end-to-end. Compared to [3], our contextual encoding method is relatively simple and can be trained end-to-end.

### 2.3. Multi-Scale Modeling for Activity Detection

To deal with activities of multiple durations in untrimmed videos, deep action proposals (DAPs) [9] use long short term memory (LSTM) to model the temporal dependencies across frames. To save computations on overlapping temporal windows, single-stream temporal action proposals (SST) [2] exploits a Gated Recurrent Unit (GRU)-based sequence encoder to model proposals of different temporal scales at each time step. The state of the Recurrent Neural Networks (RNNs) tends to saturate for long sequences, thus making the previous methods [9, 2] not appropriate for the multi-scale modeling problem. Compared to both these methods, the proposed CMS-RC3D detector constructs the temporal feature pyramid of different temporal resolutions explicitly rather than using LSTM or GRU to learn the temporal dependencies.

### 2.4. Object Detection

Recently, with the break-through results of CNNs for image classification and scene recognition [17, 27, 29, 11], generic object detectors based on CNNs, *e.g.*, the Region-based CNN (RCNN) [10], Faster RCNN [22] and other variants have been introduced [18, 6]. These methods share a similar *two-stage* pipeline (proposals followed by classification). Specifically, Faster RCNN [22] has recently achieved a balance between both detection performance and computational efficiency. To deal with small objects, the work [6] uses intermediate *conv* feature maps with fine resolution to represent small objects and learn multi-scale proposal models. Feature Pyramid Network (FPN) [18] fuses different *conv* feature maps by the lateral connection and top-down pathway, which makes the features at fine resolution more powerful and represent small objects better. These two works [6, 18] show significant improvements on small object detection due to this multi-scale fine representation.

Analogous to spatial scales in object detection, there are activities with different temporal scales in real-world videos. Inspired by [6, 18], we construct the temporal feature pyramid for a video, which consists of feature maps with different temporal resolutions to represent activities

4322

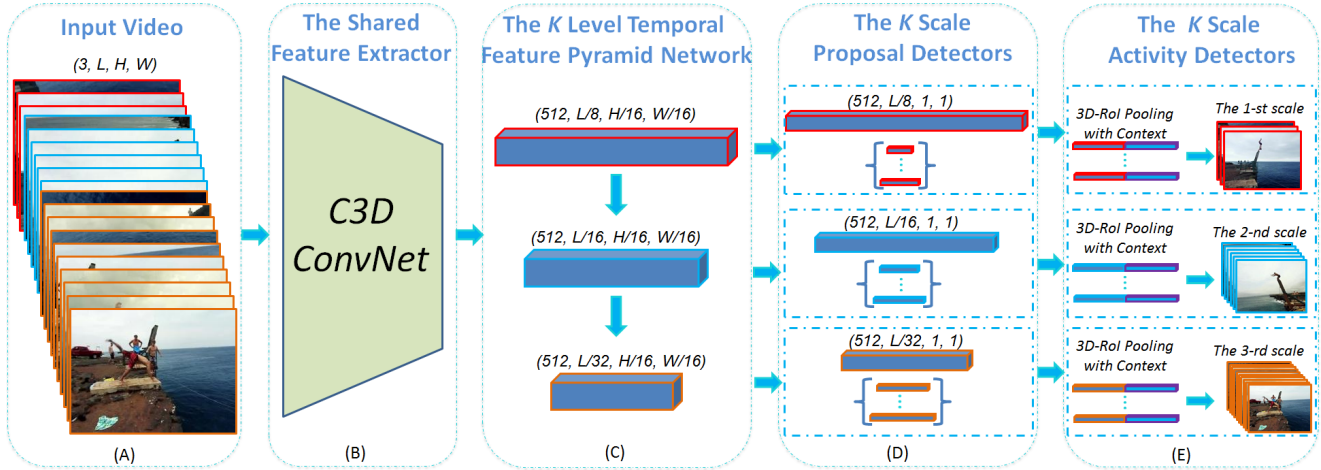

Figure 2. The pipeline of the proposed CMS-R3CD activity detector. (A) Input videos can contain activities of different temporal scales; (B) The shared feature extractor is used to extract spatio-temporal features from input videos, which can be any type of typical architectures; (C) To deal with activities of different temporal scales, the temporal feature pyramid is created, of which each level is with different temporal resolutions to represent activities of short, medium and long temporal scales simultaneously; (D) On each level of the temporal feature pyramid, an activity proposal detector is learned to detect candidate activity segments within specific temporal scale ranges; (E) 3D-RoI pooling is used to extract contextual and non-contextual features for the selected activity proposals from each level of the temporal feature pyramid. Then, the contextual and non-contextual features are concatenated and fed into the $k^{th}$ specific activity classifier, which outputs activity categories and refines activity segment boundaries.

with short, medium and long duration, leading to better fidelity in localizing activity segments.

## 3. Approach

In this section, we introduce our contextual multi-scale region convolutional 3D network (CMS-RC3D) for activity detection. As shown in Figure 2, the whole architecture consists of five components, network input, feature extraction network, temporal feature pyramid network, activity proposal and activity classification networks. In the following, we give a detailed description of each component.

### 3.1. The Network Input

The input of our network is an untrimmed video with arbitrary length, which is only limited by GPU memory. Let us denote a video $\mathbb{V}$ as a series of frames $\mathbb{V} = \{I_l\}_{l=1}^{L}$, where $I_l$ is the $l^{th}$ frame and $L$ is the total number of frames. When training, each video $\mathbb{V}$ is annotated with a set of temporal activities $\Psi = \{(t_n^{st}, t_n^{end}, c_n)\}_{n=1}^{N}$, where $N$ is the number of temporal activities in $\mathbb{V}$, $t_n^{st}$ and $t_n^{end}$ are the start-end times of the segment window for the $n^{th}$ activity. $c_n$ stands for the category and $c_n \in \{1, \cdots, C\}$, where $C$ is the number of activity categories. When inferring, we need to predict all categories of activities in a video $\mathbb{V}$, and locate the start-end times of their segment windows.

### 3.2. Feature Extraction Network

The backbone network can be any of the typical architectures, including the two-stream network [26], C3D [30], I3D [7] and P3D ResNet [21], *etc*. However, it has been shown that both spatial and temporal features are important for representing videos. A 3D ConvNet [30, 7, 21] encodes rich spatial and temporal features in a hierarchical manner, which has been experimentally demonstrated to be effective in summarizing spatio-temporal patterns from raw videos into high-level semantics. Therefore, in our current implementation, we use the shallow C3D ConvNet [30] as the backbone network to extract spatial-temporal features from a given input video buffer.

We adopt the *conv1a* to *conv5b* layers from C3D ConvNet [30] as the feature extraction network. C3D takes an input sequence $\mathbb{V}$ of RGB frames with dimension $\mathbb{R}^{3 \times L \times H \times W}$ as input, and outputs the *conv* feature maps $\mathbb{R}^{512 \times \frac{L}{8} \times \frac{H}{16} \times \frac{W}{16}}$, whereby there are 512 channel dimensions of layer *conv5b*, and 8 and 16 temporal and spatial strides. $H$ and $W$ denote the height and width of input frames, respectively. The *conv5b* feature maps are used as the shared inputs to the following temporal feature pyramid, activity proposal and classification sub-networks. Such feature sharing makes our detector efficient, since this step is the most time-consuming in the whole system.

### 3.3. Temporal Feature Pyramid Network

In unconstrained settings, activities in videos have varying temporal scales. To address this fact, many previous works (*e.g.* [25]) create different temporal scales of the input and perform several forward passes during inference. Although this strategy makes it possible to detect activities

4323

with different temporal scales, it inevitably increases the inference time. In contrast, we create a temporal feature pyramid (TFP) with different temporal strides to detect activities of different temporal scales, as shown in Figure 2 (C). More specifically, a three-level pyramid is created, which is designed to detect short, medium and long duration activities simultaneously. With this temporal feature pyramid, our detector can efficiently detect activities of all temporal scales in a single forward pass.

The temporal stride of the *conv5b* feature maps in the C3D network is $8$. To detect activities of longer durations, we add two extra branches with temporal strides of $16$ and $32$, respectively. Thus, there are $K = 3$ levels of the temporal feature pyramid and the strides of each level are $s_k \in \{8, 16, 32\}$. To create the temporal feature pyramid, down-sampling is applied on the final *conv5b* map to produce two extra level feature maps with different temporal strides. In our current implementation and for simplicity, max pooling and 3-D convolution with stride $2$ are used for down-sampling. Using this temporal feature pyramid can save a substantial amount of computational cost compared to generating multi-scale video segments, which have to pass through the backbone network several times.

### 3.4. Activity Proposal Network

Activity proposal network (APN) is a class-agnostic activity detector to find candidate activity segments for the final classification network. To detect activity proposals of different scales more efficient, the single temporal scale feature map in [33] is replaced by the multi-scale TFP of this paper and we learn an APN on each level of the TFP, as shown in Figure 2 (D). Inspired by [33], an additional 3D *conv* filters with kernel size $3 \times 3 \times 3$ operates on each level of the TFP to generate feature maps for predicting temporal proposals. The spatial dimensions of these new *conv* maps are down-sampled to produce temporal only feature maps $C_{tpn}^k \in \mathbb{R}^{512 \times \frac{L}{s_k} \times 1 \times 1}$, where $s_k$ denotes the temporal stride of the $k^{th}$ level of the TPN. Specifically, this is done by applying a 3D pooling with kernel size $1 \times \frac{H}{16} \times \frac{W}{16}$. Finally, a sub-network slides on the $C_{tpn}^k$ features densely to perform activity/non-activity binary classification and time window regression by adding two sibling $1 \times 1 \times 1$ *conv* layers on top of $C_{tpn}^k$. In order to detect activity proposals of different durations, anchors of multiple temporal scales are pre-defined, in which an anchor denotes a candidate time window associated with a scale and serves as a reference segment for activity proposals at each temporal location.

As compared to [33], each scale model in our architecture only needs to predict activities whose temporal scales lie in a limited range, rather than a single wide range. We assign anchors of specific temporal scales to each level according to training statistics. As mentioned earlier, there are $K = 3$ levels of the feature pyramid. The strides of these levels are $\{8, 16, 32\}$, respectively. As presented in Table 1, we use $\{7, 7, 11\}$ anchors of different temporal scales for the corresponding levels of the temporal feature pyramid. And the temporal scale ranges of each level are $8 - 56, 64 - 160, 192 - 512$ frames[1], respectively. Therefore, our proposal detector can handle a very wide range of temporal scales in an input video in a single shot.

Table 1. The anchor settings for each level of the temporal feature pyramid in the proposed CMS-RC3D activity detector.

| Strides | Anchor Settings | Temporal Scale Ranges |
|---|---|---|
| 8 | 1 : 7 | 8 − 56 |
| 16 | 4 : 10 | 64 − 160 |
| 32 | 6 : 16 | 192 − 512 |

**Training**: When training, we need to assign labels to the anchors based on their time intersection-over-union (tIoU) ratios with ground-truth activities. Typically, we assign an anchor with a positive label if it has the highest tIoU for a given ground-truth activity or it has a tIoU higher than $0.7$ with any ground-truth activity, and a negative label if it has a tIoU lower than $0.3$ for all ground-truth activities. All other candidates are excluded from training. For proposal segment window regression, anchors are transformed according to the corresponding ground truth activities using the coordinate transformations in [33]. The same mini-batch size of $64$ is employed for each scale model. And we fix the positive/negative ratio to 1:1.

### 3.5. Activity Classification Network

The activity classification network (ACN) is a segment-based activity detector. Due to the arbitrary length of activity proposals, 3D-Region-of-Interest (3D-RoI) pooling [33] is used to extract fixed-size features for selected proposal segments. Then, the pooled features are fed into *fc6* and *fc7* layers in the C3D network, followed by two sibling networks, which output activity categories and refine activity boundaries by classification and regression, respectively. In [33], ACN is only performed on a single-scale feature map *conv5b*. To use the TFP, we need to assign activity proposals of different temporal scales to the levels of the TFP so as to perform 3D-RoI pooling.

We propose three different strategies for assigning activity proposals to learn the ACN. In **Strategy 1** (S1), we assign all proposals to the *conv5b* feature maps (the first level of the TFP) as in [33]. However, this will not fully exploit the feature maps with multiple temporal resolutions of the TFP. In **Strategy 2** (S2), we assign proposals to the specific level of the TFP, from which proposals are generated.

---
[1]We can increase temporal scale ranges of all anchors by sampling video frames at a lower frame rate. For example, the temporal ranges are approximately $2 - 245$ seconds covering over $99\%$ training activities in ActivityNet if we sample videos at 3 frames per second.

However, the ACN for activities of long duration might not train well, since activities of long duration might be small in number. In **Strategy 3** (S3), we assign all proposals to all levels of the TFP, which takes advantage of the feature maps of multiple resolutions and avoids the issues of training ACN on less frequent activities (especially those with long duration). We validate the effectiveness of these three strategies in Section 4.3.

Our ACN can easily incorporate contextual information, which has been shown to be important for activity detection [3]. We simply double the time window of candidate proposals and use 3D-RoI pooling to extract the contextual information for the doubled time window, as shown in Figure 2 (E). To make the feature maps fit into the *fc6* layer of C3D, we reduce the number of channels in the contextual and original pooled feature maps from 512 to 256 using $3 \times 3 \times 3$ *conv* layers and concatenate them. As in [33], the new feature maps are passed through the *fc6* layer for activity prediction.

**Training**: When training the activity classifiers, an activity label needs to be assigned to each proposal generated by APN. If a proposal's highest tIoU with any ground-truth activity is greater than $0.5$, it is allocated that ground-truth activity label. A proposal is allocated the background label if its tIoU with all ground-truth activities is lower than $0.5$. For segment window regression, the start-end times are transformed according to the corresponding ground-truth activities using the coordinate transformation employed in [33]. We set the same mini-batch size of 64 with the positive/negative ratio of $1:3$ for each scale model.

### 3.6. Optimization

To train the proposal and classification networks, we need to optimize both classification and regression tasks jointly. Following conventions in the field, the softmax loss function and the smooth L1 loss function are used for classification and time window regression, respectively. Specifically, the objective loss function is defined by:

$$L(\mathbf{y}, \mathbf{t}) = \sum_{k=1}^{K} \gamma_k \sum_{i=1}^{N} (L_{cls}(y_{ki}, y_{ki}^*) + \lambda_k y_{ki}^* L_{loc}(\mathbf{t}_{ki}, \mathbf{t}_{ki}^*))$$
(1)

where $\mathbf{y} = [y_{1i}, ... y_{1N}, ..., y_{K1}, ..., y_{KN}]$ and $\mathbf{t} = [\mathbf{p}_{1i}, ... \mathbf{t}_{1N}, ..., \mathbf{t}_{K1}, ..., \mathbf{t}_{KN}]$ denote the vectors of predicted labels and time windows of the corresponding anchors or proposals, respectively. $\mathbf{y}^* = [y_{1i}^*, ... y_{1N}^*, ..., y_{K1}^*, ..., y_{KN}^*]$ and $\mathbf{t}^* = [\mathbf{t}_{1i}^*, ... \mathbf{t}_{1N}^*, ..., \mathbf{t}_{K1}^*, ..., \mathbf{t}_{KN}^*]$ denote the vectors of ground-truth labels and time windows. $N$ denotes the mini-batch size, while $\gamma_k$ balances the importance of models at different branches. In our experiments, each $\gamma_k$ is set to 1, which means that all $K$ sub-models have the same importance. This can be changed to reflect the duration statistics of activities in the training data. $\lambda_k$ is the tradeoff parameter between classification and localization. In our experiments, each $\lambda_k$ is set to 1.

### 3.7. Inference

Since our method falls into the detection by classification category, there are two main steps for activity detection in our model. First, APN scores all anchors and predicts start-end segment window offsets. Then, non-maximum suppression (NMS) with tIoU threshold $0.7$ is applied to all anchors of all temporal scales to obtain final proposal candidates. Second, the selected proposals are classified into specific activities and segment boundaries are further refined by ACN. Finally, NMS with a lower tIoU threshold $0.4$ is applied to predicted activities to get the final results.

## 4. Experiments

In this section, we experimentally validate our proposed CMS-RC3D detector on two large-scale action detection benchmark datasets: THUMOS14 [15] and ActivityNet [5]. Firstly, we introduce these datasets and other experimental settings and then investigate the impact of different components via a set of ablation studies on the ActivityNet validation set. Secondly, we evaluate our CMS-RC3D detector on both these public benchmarks, while comparing the performance against other state-of-the-art approaches.

### 4.1. Evaluation Datasets

**THUMOS14** [15] contains 1010 videos for validation and 1574 videos for testing. These videos contain annotated action instances from 20 different sport activities. This dataset is particularly challenging as there are many activity instances of very short duration in very long videos. Since the training set is trimmed, we resort to training our models on its validation set and evaluate them on the testing set. In summary, there are about 200 untrimmed videos from 20 action classes for training and 213 videos for testing.

**ActivityNet** [5] is a recently released large-scale activity detection benchmark. There are two versions, and we use the latest version 1.3, which consists of 10024, 4926 and 5044 videos in the training, validation and testing subsets respectively. ActivityNet contains 200 different types of activities. Many videos in this dataset contain activity instances of a single class covering most of the duration of the video. Compared to THUMOS14 [15], ActivityNet has much more diversity both in terms of the number of activity categories and the number of videos. Since the ground-truth annotations of test videos are not public, following traditional evaluation practices on this dataset and for fair comparison, we use the evaluation subset for ablation studies.

**Evaluation Metrics.** Each dataset has its own convention of performance metrics. Following their conventions,

we report the mean average precision (mAP) at different tIoU thresholds. On THUMOS14, we report the detection results at tIoU thresholds $\{0.1, 0.2, 0.3, 0.4, 0.5\}$. Following traditional evaluation practises, the mAP at $0.5$ tIoU is used for comparing final results from different methods. On ActivityNet, we use the standard evaluation protocol and report the mAP at different tIoU thresholds, *i.e.* $0.5$, $0.75$, $0.95$, and the average of mAP values with tIoU thresholds $[0.5 : 0.05 : 0.95]$ for comparing different methods.

### 4.2. Implementation Details

**Training setup.** When training the CMS-RC3D detector, each mini-batch is constructed from one video, chosen uniformly at random from the training dataset. Each frame in a video is resized to $172 \times 128$ (width $\times$ height) pixels, and we randomly crop regions of $112 \times 112$ from each frame. To fit each batch into GPU memory, we create a video buffer of $768$ frames. The buffer is generated by sliding from the beginning of the video to the end and from the end of the video to the beginning to increase the amount of training data, which can be seen as a data augmentation strategy. We also employ data augmentation by horizontally flipping all frames in a batch with a probability of $0.5$.

**Hyper-parameters.** The weights of the filters of all new added layers are initialized by randomly drawing from a zero-mean Gaussian distribution with standard deviation $0.01$. Biases are initialized at $0.1$. All other layers are initialized from the pre-trained C3D model. The learning rate is initially set to $10^{-4}$ and then reduced by a factor of $10$ after every $100k$ iterations. Training is terminated after a maximum of $150k$ iterations. We also use a momentum of $0.9$ and a weight decay of $5 \times 10^{-4}$. Our system is implemented in Caffe [14] and its source code will be made publicly available.

### 4.3. Ablation study

**Multi-Scales vs Single-Scale.** The main strength of our multi-scale model is that it uses a *conv* feature map pyramid with different temporal resolutions for activity detection of different temporal scales. Compared to our baseline method RC3D [33], our MS(MAX)(S1) and MS(CONV)(S1) use the temporal feature pyramid for proposal generation and assign proposals on the *conv5* layer for 3D ROI-Pooling. Detection results from our multi-scale models register an improvement of $2.2\%$ and $2.4\%$ in the average mAP over those from the single-scale model (refer to the $1^{st}$, $2^{nd}$ and $3^{rd}$ rows of Table 2). We also present the proposal detection results in Table 3, which shows better results of multi-scale models over the single scale model. These highlight the motivation of our method.

**How to Construct the Temporal Feature Pyramid?** As we discuss in Section 3.3, we can employ max pooling or

Table 2. Ablation study of each component in the CMS-RC3D framework. Action detection results are measured by mean average precision (mAP) @0.5, mAP@0.75, mAP@0.95 and the average mAP of tIoU thresholds from $0.5 : 0.05 : 0.95$ on ActivityNet validation dataset. MS denotes the multi-scale model. MAX and CONV denote different methods to construct the temporal feature pyramid. S1, S2 and S3 denote different strategies for assigning proposals. CTX denotes contextual modeling. The same annotations are used in Table 3.

| Method | 0.5 | 0.75 | 0.95 | Average |
|---|---|---|---|---|
| RC3D [33] | 26.33 | 10.46 | 1.25 | 12.71 |
| MS(MAX)(S1) | 27.65 | 13.93 | 1.12 | 14.91 |
| MS(CONV)(S1) | 28.01 | 13.80 | 1.20 | 15.12 |
| MS(MAX)(S1)(CTX) | 31.81 | 17.05 | 1.06 | 17.58 |
| MS(CONV)(S1)(CTX) | 32.57 | 16.92 | 1.07 | 17.89 |
| MS(CONV)(S2)(CTX) | 31.89 | 17.23 | 1.16 | 17.72 |
| MS(CONV)(S3)(CTX) | 32.92 | 18.36 | 1.13 | 18.46 |

Table 3. The activity proposal detection results generated by R-C3D and our multi-scale models.

| Method | Prop. | AR |
|---|---|---|
| RC3D [33] | 100 | 56.0 |
| MS(MAX)(S1) | 100 | 56.6 |
| MS(CONV)(S1) | 100 | 56.8 |

*conv* layers to construct the TFP. As shown in the $2^{nd}$ and $3^{rd}$ rows of Table 2, we observe that the MS(CONV)(S1) model (*i.e.* a model using *conv* for the TFP shows slightly better performance than the MS(MAX)(S1) model (*i.e.* a model using max pooling for the TFP). The improvement can be also observed when incorporating contextual information (refer to the $4^{th}$ and $5^{th}$ rows of Table 2).

**Contextual vs. Non-contextual Modeling.** Here, we evaluate the impact of temporal context information on detection performance. To incorporate contextual information, We simply double the time window of activity proposals and use 3D-RoI pooling to extract contextual features. Table 2 lists the performance of models with and without contextual modeling. From these results, we conclude that contextual modeling can significantly improve detection performance (about $2.5\%$ improvement in the average mAP; refer to the $2^{nd}$ and $4^{th}$, the $3^{rd}$ and $5^{th}$ rows of Table 2). Activities are continuous in nature and temporal context gives information on when an activity might begin and end. Therefore, contextual modeling can lead to better activity localization, thus, improve the detector's performance.

**Strategies For Assigning Proposals.** To study the effect of assigning proposals for training the activity classifier discussed in Section 3.5, we evaluate the detection performances of different assigning strategies, which are presented in the $5^{th}$-$7^{th}$ rows in Table 2. From the comparisons, the detector with Strategy 3 achieves the best performance.

Table 4. Activity detection results on THUMOS14 testing dataset, measured by the mean average precision (mAP) of different tIoU thresholds $\alpha$.

| Method | 0.1 | 0.2 | 0.3 | 0.4 | 0.5 |
|---|---|---|---|---|---|
| Karaman *et al.* [16] | 4.6 | 3.4 | 2.1 | 1.4 | 0.9 |
| Wang *et al.* [31] | 18.2 | 17.0 | 14.0 | 11.7 | 8.3 |
| Oneata *et al.* [20] | 36.6 | 33.6 | 27.0 | 20.8 | 14.4 |
| SparseProp [4] | - | - | - | - | 13.5 |
| DAPs [9] | - | - | - | - | 13.9 |
| SLM [23] | 39.7 | 35.7 | 30.0 | 23.2 | 15.2 |
| FG [35] | 48.9 | 44.0 | 36.0 | 26.4 | 17.1 |
| PSDF [36] | 51.4 | 42.6 | 33.6 | 26.1 | 18.8 |
| S-CNN [25] | 47.7 | 43.5 | 36.3 | 28.7 | 19.0 |
| CDC [24] | - | - | 40.1 | 29.4 | 23.3 |
| TCN [8] | - | - | - | 33.3 | 25.6 |
| RC3D [33] | 54.5 | 51.5 | 44.8 | 35.6 | 28.9 |
| SS-TAD [1] | - | - | - | 45.7 | 29.2 |
| SSN [37] | 66.0 | 59.4 | 51.9 | 41.0 | 29.8 |
| Our RC3D | 57.4 | 54.9 | 51.1 | 43.1 | 35.8 |
| CMS-RC3D | 61.6 | 59.3 | 54.7 | 48.2 | 40.0 |

Table 5. Activity detection results on ActivityNet v1.3 testing dataset. The performances are measured by mean average precision (mAP) at different tIoU thresholds $\alpha$ and the average mAP of tIoU thresholds from $0.5 : 0.05 : 0.95$.

| Method | 0.5 | 0.75 | 0.95 | Average |
|---|---|---|---|---|
| RC3D [33] | 26.45 | 11.47 | 1.69 | 13.33 |
| MSN [28] | 28.67 | 17.78 | 2.88 | 17.68 |
| TCN [8] | 37.49 | 23.47 | 4.47 | 23.58 |
| SSN [37] | 43.26 | 28.70 | 5.63 | 28.28 |
| CMS-RC3D | 32.79 | 18.39 | 1.24 | 18.68 |

In Strategy 3, all proposals are assigned to each level of the TFP, which takes advantage of the temporal feature pyramid of different temporal resolutions and avoids the issues of training the activity classifiers on less frequent activities. This demonstrates that the advantage of using the TFP when training the activity classifiers.

From the above comparisons, we conclude that modeling activities with the temporal feature pyramid of different temporal resolutions and encoding temporal contextual information can increase the activity detection performance significantly. In what follows, the CMS-RC3D detector is chosen to be the MS(CONV)(S3)(CTX) model, which learns the activity detector via using *conv* layers to construct the temporal feature pyramid, Strategy 3 for assigning activity proposals and contextual modeling.

### 4.4. Evaluation on THUMOS14 [15]

In Table 4, we summarize the results of our CMS-RC3D detector for the task of temporal activity detection on THUMOS14 in comparison to existing state-of-the-art detectors. We re-train the origin RC3D model with better hyperparameter fine-tuning as our baseline model (*i.e.* akin to the single-scale version of our model), and we achieve better results for the original RC3D compared to the results reported in [33] (about $7.0\%$ improvement @mAP $0.5$; refer to the $12^{th}$ and $15^{th}$ rows of Table 4). The CMS-RC3D shows about $4.0\%$ improvement @mAP $0.5$ over our RC3D model (refer to the last two rows of Table 4), which indicates the importance of exploiting the temporal feature pyramid and context information. Moreover, we also observe that CMS-RC3D registers superior performance over prior state-of-the-art approaches at high tIoU thresholds (about $10.0\%$ improvement @mAP $0.5$ over the $2^{nd}$ best method except our RC3D model), thus, indicating that the CMS-RC3D detector can more precisely localize activity intervals in untrimmed video sequences. This result provides even more justification of the temporal feature pyramid for multi-scale activity detection. Some qualitative detection results on THUMOS14 are shown in Figure 3(a).

### 4.5. Evaluation on ActivityNet [5]

To measure the performance of CMS-RC3D on the ActivityNet v1.3 testing dataset, we submit our prediction results to the publicly available evaluation server maintained by the ActivityNet organizers. As shown in Table 5, we compare CMS-RC3D with state-of-the-art methods [33, 28, 8, 37] published recently. From the comparisons, CMS-RC3D shows a significant improvement (about $5.0\%$ in the average mAP of tIoU thresholds from $0.5 : 0.05 : 0.95$) over our baseline method RC3D [33], which demonstrates the motivation of our method. CMS-RC3D shows inferior performance over TCN [8] and SSN [37]. However, both these methods use the two-stream network [26] as their feature extractor, which has been shown to be more discriminative than the C3D extractor deployed in our model. Compared to the much deeper two-stream network, which uses RGB and optical flow information, C3D only uses low resolution RGB information. We expect that the incorporation of flow in CMS-RC3D to further improve detection results as attested by previous works that make use of this extra information (about $4.0\%$ in average mAP [8]). Moreover, our model is agnostic of the exact manifestation of the feature extractor and it can be changed into other networks, *e.g.*, I3D [7] or P3D ResNet [21]. Some qualitative detection results on ActivityNet are shown in Figure 3(b).

## 5. Conclusion

In this paper, we propose a contextual multi-scale region convolutional C3D network for activity detection. In our CMS-RC3D, each scale model uses *conv* feature maps of specific temporal resolution to represent activities within specific temporal scale ranges. More importantly, the w-

4327

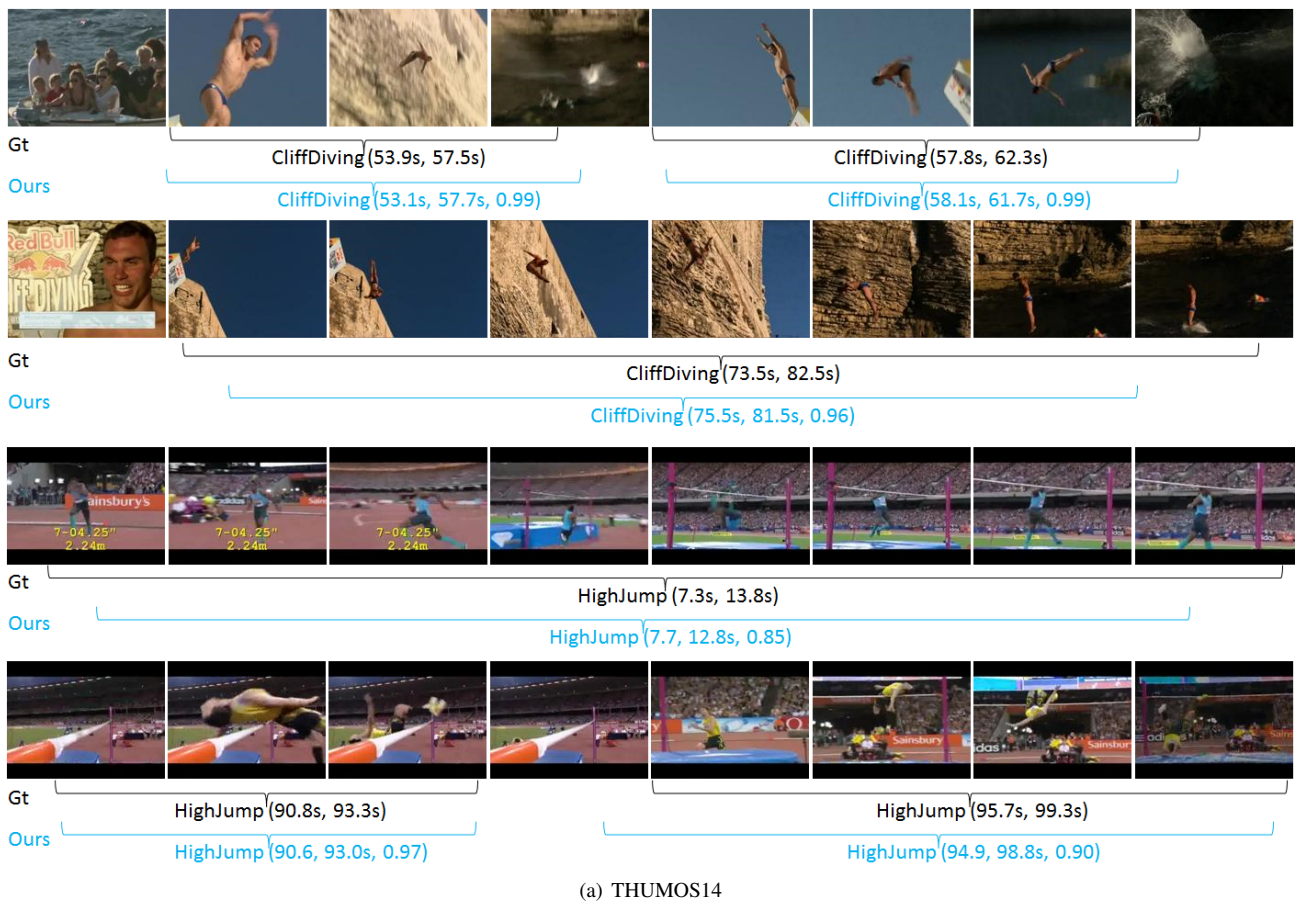

(a) THUMOS14

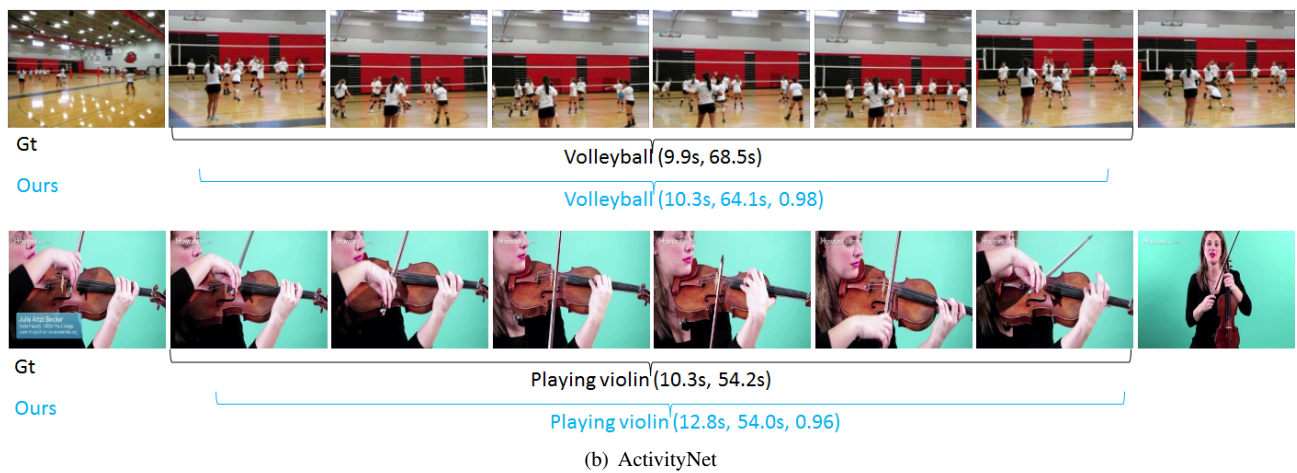

(b) ActivityNet

Figure 3. Qualitative visualization of the predicted activities by CMS-RC3D (best viewed in color). Figure 3(a) and 3(b) show results for two videos each in THUMOS14 and ActivityNet. The ground-truth and predicted activity segments are marked in black and light blue respectively. Corresponding start-end times and confidence scores are shown inside brackets. From the results, we see that the CMS-RC3D detector can deal with activities of different temporal scales.

hole detector can detect activities within all temporal ranges in a single shot, which makes it computationally efficient. CMS-RC3D is evaluated on two public activity detection benchmarks, THUMOS14 and ActivityNet, and achieves superior or at least comparable performance compared to other state-of-the-art detectors. Moreover, deep feature extractors can be used in the CMS-RC3D detector to further improve the performance. We will exploit various feature extractors and see their influence on activity prediction.